\def\year{2016}\relax
\pdfoutput=1
\documentclass[letterpaper]{article}
\usepackage{xcolor}
\usepackage{aaai16}
\usepackage{times}
\usepackage{helvet}
\usepackage{courier}
\usepackage{mathrsfs}
\usepackage{amssymb,graphicx,amsfonts}
\usepackage[cmex10]{amsmath}
\usepackage{float}
\usepackage{url}

\newcommand{\M}{\Phi} 
\newcommand{\p}{\rho} 

\newcommand{\x}{x} 

\newcommand{\Dtrn}{X_{train}}
\newcommand{\Dtst}{X_{test}}
\newcommand{\Dvld}{X_{valid}}

\begin{document}
\title{High Dimensional Human Guided Machine Learning}
\author{Eric Holloway, Robert Marks II \\
  Dept. of Electrical \& Computer Engineering \\
  Baylor University \\
  Waco, Texas\\
email: first\_last @ baylor.edu}
\maketitle
\begin{abstract}
  Have you ever looked at a machine learning classification model and thought, I could have made that?
  Well, that is what we test in this project, comparing XGBoost trained on human engineered features to training directly on data.
  The human engineered features do not outperform XGBoost trained directly on the data, but they are comparable.
  This project contributes a novel method for utilizing human created classification models on high dimensional datasets.
\end{abstract}

\section{Why Human Guided?}

In the artificial intelligence, machine learning, and human computation fields there is little research into the effectiveness of human generated models.
One research project is human guided simple search \cite{anderson2000human}, and tabu search \cite{klau2002human}.
 Humans outperform the state of the art algorithms when solving complex visual problems, such as the travelling salesman problem \cite{krolak1971man,dry2006human,acuna2010people}.
Numerous machine learning algorithms are NP-Complete or harder, such as the set cover machine (SCM) \cite{marchand2003set}.
Breakthroughs have been achieved by including humans-in-the-loop for hard optimization and combinatorial problems, \cite{le2014human} and \cite{khatib2011crystal}.
With these promising results there is need for further investigation into human guided machine learning.

Machine learning algorithms typically work with high dimensional datasets, which a human cannot visualize in entirity.
But the high dimensionality of a dataset is not an insurmountable obstacle to effectively using a human-in-the-loop.

\section{Approach and Implementation}
In this project we use a dimension subset approach to test out human effectiveness in creating classification models.
Instead of having a human attempt high dimensional visualization, we have humans design models on pairs of dimensions.
These models are then used to transform the dataset into a feature space.

XGBoost \cite{chen2016xgboost}, short for eXtreme Gradient Boosting, is a popular machine learning library that has been used to win multiple Kaggle competitions.
An XGBoost model is trained on the transformed data, and the results are compared to training XGBoost on the untransformed data.
We are not restricted to only using XGBoost, other machine learning approaches also work and we have tested linear perceptrons, linear regression and support vector machines.

\begin{figure}
  \centering
  \includegraphics[width=2.5in]{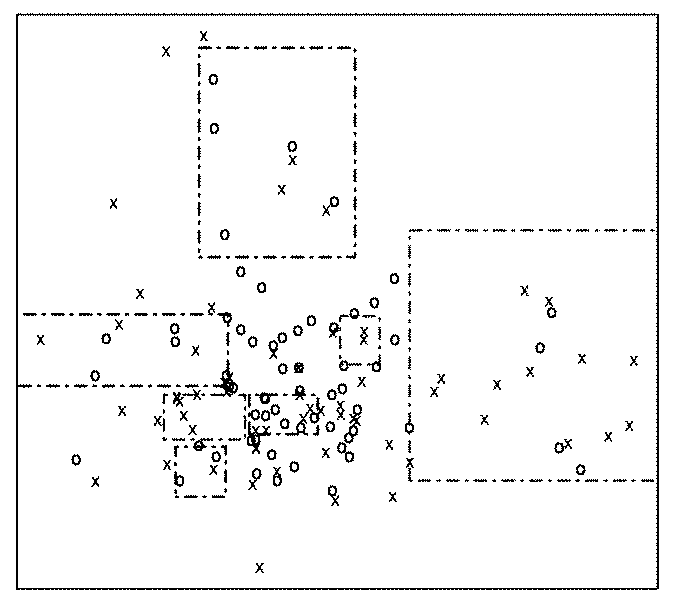}
  \caption{Example of polygons drawn by worker.}
  \label{fig:best_act_model}
\end{figure}

The following process is used to create each model.
\begin{enumerate}
\item A pair of dimensions are selected and the training dataset ($\Dtrn$) centered and normalized for those dimensions.
  The training dataset contains about 100 samples.
  The pairs are selected based on low correlation between the two dimensions.
  If there is low correlation, then it is easier to identify clusters of data points.
  Not all dimensions in a dataset are used by the workers.
\item The worker is given a scatterplot of the two dimensions and proceeds to draw polygons ($\p$) to separate the data into classification regions.
  Each polygon classifies a sample to one class ($\p_{class}$).
  For simplicity, the polygon is a rectangle, making the models similar to those produced by SCM \shortcite{marchand2003set}.
\item The collection of polygons drawn by the worker on a pair of dimensions is a single model ($\M$).
  An example of a model is shown in Figure \ref{fig:best_act_model}.
\item The model is evaluated on a test dataset ($\Dtst$) producing an accuracy score for the classification regions ($\M_{acc}$), see Equation \ref{eqn:acc}.
  Only samples contained by a polygon ($\Dvld^\p$) in the model ($\Dvld^{\M}$) contribute to the accuracy score.
  The test dataset contains about 200 samples.
\end{enumerate}
\begin{align}
  \label{eqn:acc}
  \M_{acc} = \frac{\sum\limits_{\p \in \M}\sum\limits_{\x \in \Dvld^\p} [\x_{class} = \p_{class}]}{|\Dvld^{\M}|}
\end{align}

The sample transformation function is shown in Equation \ref{eqn:ftr}, which is a weighted sum of model polygons containing the sample.
\begin{align}
  \label{eqn:ftr}
  f(\x, \M) = [\x \in \M] * \M_{acc}
\end{align}
Then, for $M$ samples and $N$ models we have the following $M \times N$ feature matrix.
\[
\begin{bmatrix}
  f(x_1,\M_1) & f(x_1,\M_2) & \ldots & f(x_1,\M_N) \\
  f(x_2,\M_1) & f(x_2,\M_2) & \ldots & f(x_2,\M_N) \\
  \vdots & \vdots & \ddots & \vdots \\ 
  f(x_M,\M_1) & f(x_M,\M_2) & \ldots & f(x_M,\M_N) \\
\end{bmatrix}
\]

XGBoost is trained on a subset $M'$ of the $M$ samples, and then used to classify the remaining samples.
To perform a fair comparison, only the dimensions used by the workers are included in the untransformed samples.
For example, if the dataset has $D$ dimensions, but only $D'$ dimensions are used, the XGBoost model is trained on an $M' \times D'$ matrix.
Thus, one  XGBoost model is trained on the untransformed samples ($M' \times D'$ data matrix), and another on the transformed samples ($M' \times N$ feature matrix).

The Amazon online service Mechanical Turk (AMT) is used to gather human produced models.
\begin{enumerate}
\item The AMT job directs the worker to a website where they can perform the classification task.
\item A scatterplot shows $\Dtrn$ plotted according to the randomly chosen dimension pair, and the worker draws boxes on the scatterplot.
\item A progress bar gives feedback on the accuracy of the model.
  Accuracy is calculated on a validation dataset $\Dvld$.
  The validation dataset contains about 100 samples.
  Only models that achieve an accuracy above 50\% are accepted, to provide quality control.
\item Once the model has been accepted, the website gives the worker a job completion code.
\item Back at the AMT job posting, the worker submits the code for payment.
\end{enumerate}

We use five datasets with binary classification tasks.  Datasets consist of one synthetic clustering task, and the rest are real world datasets from Kaggle.  Most of the datasets are highly unbalanced, so we balance the datasets to have an equal number of both classes.  Additionally, with the exception of the synthetic dataset, the dimensions consist of both nominal and continuous variables.  A summary of the datasets is in Table \ref{tbl:datasets}, and the dataset sources are the following.
\begin{itemize}
\item Madelon (Mad.) \cite{guyon2004result}
\item Carvana (Car.)  \cite{carvana11:_dont}
\item Homesite (Home.) \cite{homesite15:_homes_quote_conver}
\item Melbourne (Mel.) \cite{melbourne10:_predic_grant_applic}
\item Credit \cite{kaggle11:_give_me_some_credit}
\end{itemize}

\begin{table}
  \begin{tabular}{|r||c|c|c|c|}
    \hline
    Name  & Nom. & Int. & Cont. & Note \\ \hline
    Mad. & 0 & 500 & 0 & hyper-XOR problem \\
    Car. & 18 & 0 & 14 & car auction \\
    Home. & 295 & 0 & 1 & real estate \\ 
    Mel.  & 178 & 61 & 11 & grant applications \\ 
    Credit  & 0 & 6 & 4 & credit risks\\
    \hline													
  \end{tabular}
  \caption{Datasets and their characteristics. Nom = nominal. Int = integer. Cont = continuous.}
  \label{tbl:datasets}
\end{table}

\section{Results and Conclusion}

Table \ref{tbl:results} demonstrates the results from training XGBoost directly on the data, as well as on the features generated by the AMT workers.
XGBoost's model is parameterized by cross validation.
The parameters are learning rate (0.01, 0.05, 0.1, 0.3), max tree depth (2, 5, 10, 15), and number of rounds (50, 100, 200, 400, 800).

\begin{table}
  \begin{tabular}{|r||c|c|c|c|c|c|}
    \hline
    Name  & M' & M-M' & D' & \emph{Data} & N & \emph{Features} \\ \hline
    Mad. & 2000 & 600 & 73 & 0.650 & 320 & 0.655 \\
    Car. & 2000 & 2230 & 7 & 0.521 & 194 & 0.481 \\
    Home. & 2000 & 1806 & 43 & 0.795 & 194 & 0.723 \\ 
    Mel. & 2000 & 2500 & 9 & 0.542 & 64 & 0.512 \\ 
    Credit & 2000 & 18052 & 8 & 0.762 & 156 & 0.717\\
    \hline													
  \end{tabular}
  \caption{Accuracy results of training XGBoost directly on \emph{data}, and on \emph{features} produced by AMT workers.
    M is the total number of samples.
    M' is the number of samples in the training dataset.
    M-M' is the number of samples in the test dataset.
    D' is the number of dimensions used by the workers.
    N is the number of models the workers created and the number of features generated.}
  \label{tbl:results}
\end{table}

We've shown that human guided machine learning can be crowd sourced through workers drawing polygons on scatterplots.
These models do not outperform standard algorithmic approaches, but are comparable.
The contribution of this project is human model creation on high dimensional datasets.

Future research will discover if and when human produced models outperform purely algorithmic approaches.
In this research, human produced models did not outperform algorithmic approaches likely due to loss of information.
Transforming the data using the models reduces the data granularity.
A way ahead is to find a way to preserve granularity while using the human produced models.
\section{Acknowledgements}
The researchers thank the AMT workers who contributed their valuable insight.
\bibliography{references}
\bibliographystyle{aaai}
\end{document}